\documentclass[10pt,twocolumn,letterpaper]{article}

\usepackage[pagenumbers]{cvpr} % To force page numbers, e.g. for an arXiv version

\definecolor{cvprblue}{rgb}{0.21,0.49,0.74}
\usepackage[pagebackref,breaklinks,colorlinks,allcolors=cvprblue]{hyperref}
\usepackage{algorithm}
\usepackage{algorithmic}
\usepackage{multicol, multirow}
\usepackage[table]{xcolor}
\usepackage{pifont}
\usepackage{graphicx}
\usepackage{amsmath}
\usepackage{amssymb}
\usepackage[utf8]{inputenc}
\usepackage{tcolorbox}
\usepackage{listings}
\usepackage{xcolor}
\usepackage{bibentry}
\usepackage{geometry}
\definecolor{lightgray}{gray}{0.95}

\def\paperID{***} % *** Enter the Paper ID here
\def\confName{CVPR}
\def\confYear{2026}

\title{Towards Robust Text-to-Image Person Retrieval: Multi-View Reformulation for Semantic Compensation}

\author{
Chao Yuan$^{*}$, Yujian Zhao$^{*}$, Haoxuan Xu, Guanglin Niu$^{\dagger}$ \\
Beihang University \quad \\
{\tt\small yuanc3666@gmail.com}, {\tt\small beihangngl@buaa.edu.cn} \\
}
\begin{document}
\maketitle
\footnotetext{$\dagger$ Corresponding author. * Equal contribution.}

\begin{abstract}
In text-to-image person retrieval tasks, the diversity of natural language expressions and the implicitness of visual semantics often lead to the problem of Expression Drift, where semantically equivalent texts exhibit significant feature discrepancies in the embedding space due to phrasing variations, thereby degrading the robustness of image-text alignment. This paper proposes a semantic compensation framework (MVR) driven by Large Language Models (LLMs), which enhances cross-modal representation consistency through multi-view semantic reformulation and feature compensation. The core methodology comprises three components: \ding{192} \textbf{M}ulti-\textbf{V}iew \textbf{R}eformulation (\textbf{MVR}): A dual-branch prompting strategy combines key feature guidance (extracting visually critical components via feature similarity) and diversity-aware rewriting to generate semantically equivalent yet distributionally diverse textual variants; \ding{193} Textual Feature Robustness Enhancement: A training-free latent space compensation mechanism suppresses noise interference through multi-view feature mean-pooling and residual connections, effectively capturing "Semantic Echoes"; \ding{194} Visual Semantic Compensation: VLM generates multi-perspective image descriptions, which are further enhanced through shared text reformulation to address visual semantic gaps. Experiments demonstrate that our method can improve the accuracy of the original model well without training and performs SOTA on three text-to-image person retrieval datasets.
\end{abstract}

\begin{figure}
\centering
\includegraphics[width=\linewidth]{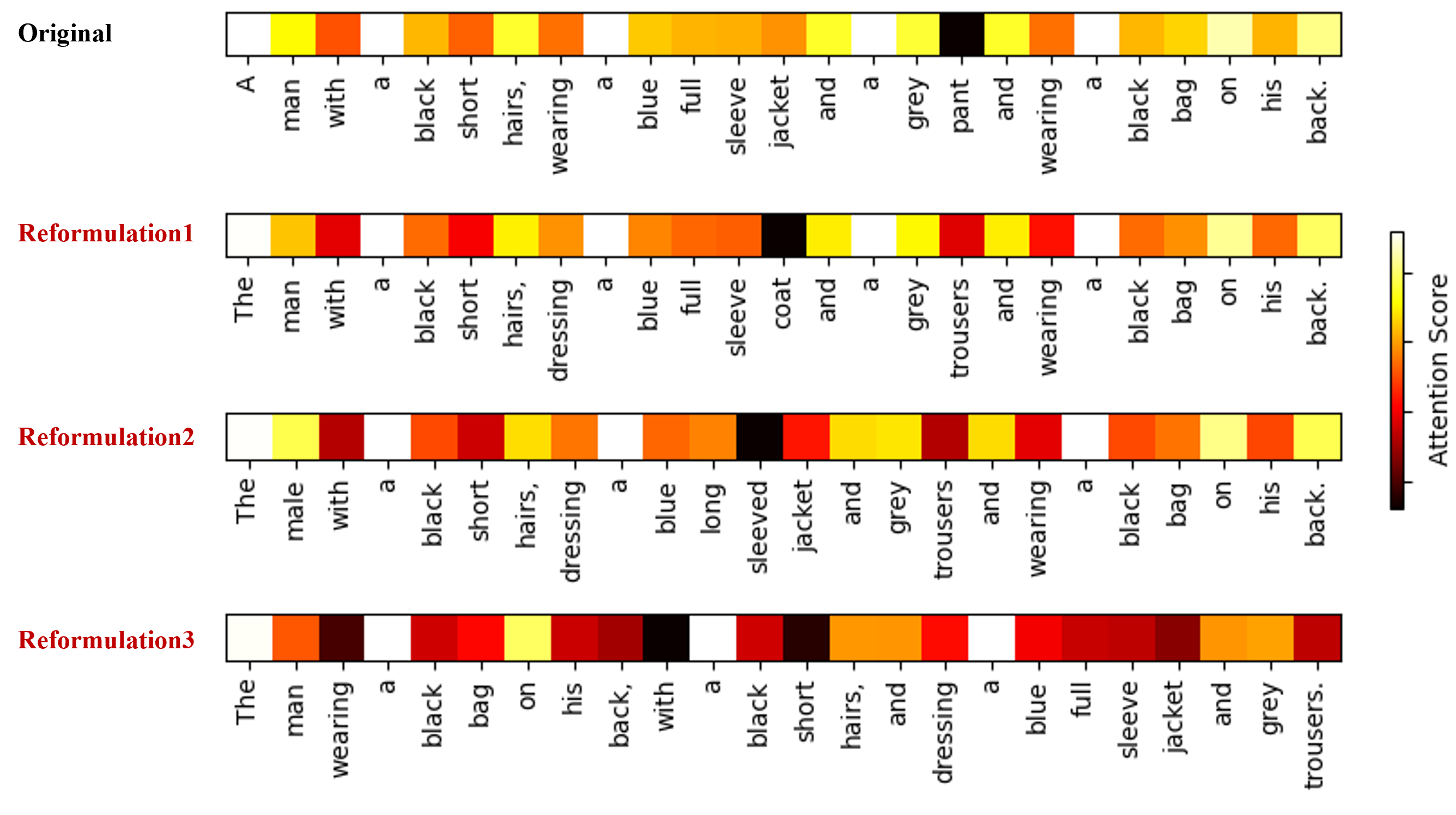}
\caption{Cosine similarity visualization between word vectors and sentences with same meaning but different words/orders. It shows that sentences with the same meaning may pay more attention to different words.}
\label{fig:intro}
\end{figure}

\section{Introduction}
Text-to-image person retrieval \cite{CUHK-PEDES, ding2021semantically, zhu2021dssl} targets the challenging task of retrieving person images from a large-scale gallery based on natural language descriptions. As a cross-modal retrieval problem, it demands precise semantic alignment between heterogeneous modalities (i.e., vision and language). In recent years, this task has attracted attention due to its broad applicability, ranging from personal photo album search to intelligent surveillance and public security. The ability to bridge text with visual person representations holds transformative potential across numerous real-world scenarios.

% stands as a foundational task bridging computer vision and natural language processing. 

Recent advances in contrastive learning-based vision-language pretrained model (VLM)~\cite{radford2021learning, li2022blip} have enabled impressive progress by learning joint embeddings from large-scale image-text pairs. Building upon these foundations, many state-of-the-art text-to-image person retrieval methods \cite{HAM, zuo2024plip, hu2024empowering} leverage pretrained VLM to bridge the semantic gap between textual and visual features. However, these methods still face significant challenges when confronted with the inherent diversity of natural language expressions and the ambiguity of visual semantics. On the textual side, the richness of human language (e.g., synonym substitution, syntactic variation) often leads to unintended feature dispersion of semantically equivalent texts in the embedding space. On the visual side, implicit semantics in images (e.g., abstract attributes) are rarely fully captured, resulting in semantic gaps during cross-modal alignment. Achieving robust semantic consistency across modalities remains a critical bottleneck for retrieval performance.

Our motivation stems from a fundamental observation that natural language is inherently diverse and compositional: semantically equivalent statements can be expressed in vastly different ways. Similarly, visual content may contain abstract semantics that can not be precisely captured by raw pixel-based features alone. However, existing methods typically encode each modality as a static point in the embedding space, overlooking the variability and ambiguity present in real-world scenarios. As illustrated in Fig.~\ref{fig:intro}, different textual descriptions of the same person may lead the model to focus on different words, which may get different feature representations. In some cases, some descriptions may lay high attention to non-critical words while diluting the contribution of critical ones, (e.g., the Reformulation 3 in Fig.\ref{fig:intro}), resulting in retrieval failures. This highlights the need for more robust and semantically consistent representations across modalities.

To address these issues, we propose a simple yet effective \textbf{training-free semantic compensation framework} (MVR) that enhances both textual and visual representations through multi-view reformulation and cross-modal semantic compensation. Our method is built upon the following two intuitions: (1) semantically robust features can be distilled from multiple reformulations of the same input, and (2) language can serve as a powerful semantic bridge to enhance vision representations. We design two parallel compensation modules: one for text and one for vision. 
In the textual branch, we leverage large language models (LLMs) in a collaborative fashion to construct multi-view reformulations of a query and integrate their features via residual mean pooling. In the visual branch, we translate image content into natural language using a pre-trained VLM, augment the textual descriptions via multi-view LLM reformulations, and project them back to semantic space to compensate the original visual embedding. 

These reformulations are encoded by frozen text encoders, and their features are aggregated in latent space to form a semantically enriched representation. Together, these modules significantly improve representation robustness under semantic shift and input noise. Note that this entire process requires no training, fine-tuning or additional supervision.

To summarize, our main contributions are as follows:
\begin{itemize}
\item We identify and systematically address the vulnerability of current text retrieval systems to \textbf{expression variance and semantic drift}.
\item We introduce a novel training-free \textbf{M}ulti-\textbf{V}iew \textbf{R}eformulation (\textbf{MVR}) framework that enhances both text and visual features through multi-view semantic compensation.
\item We propose a unified formulation that distills semantically invariant signals via residual composition of frozen encoder features, without training.
\item Extensive experiments on standard text-to-image person retrieval benchmarks demonstrate that our method significantly improves robustness of features, but the extra computational overhead will not affect its \textbf{real-time performance}.
\end{itemize}

\begin{figure*}
\centering
\includegraphics[width=\textwidth]{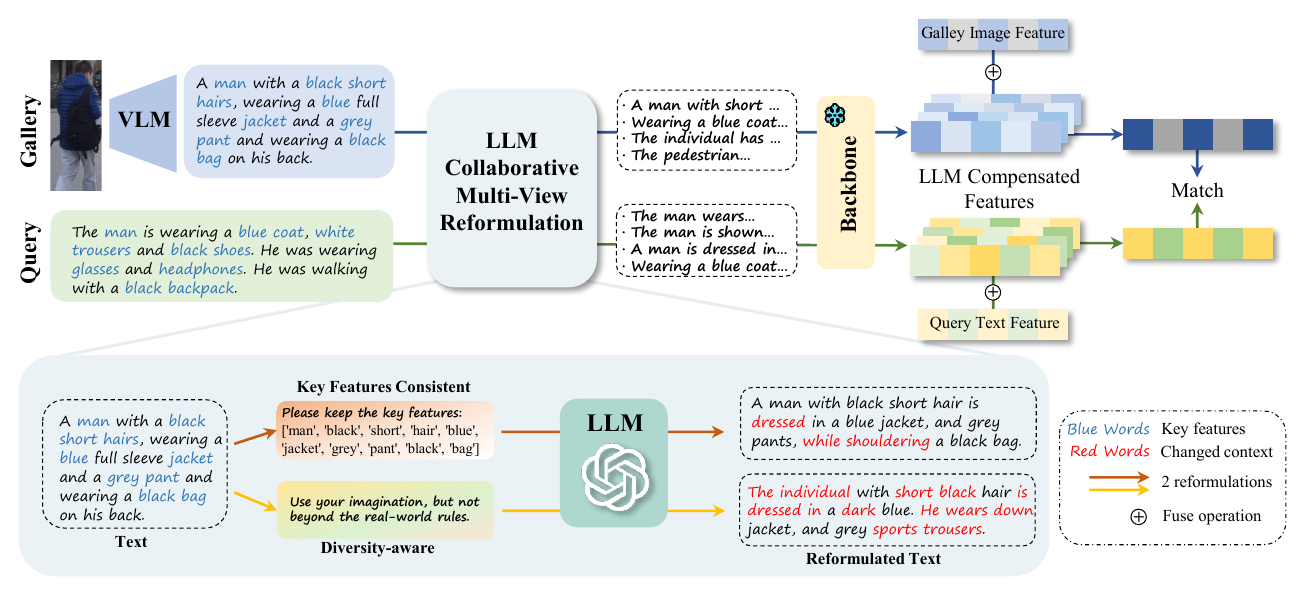}
\caption{Overview of the proposed framework (MVR) for the text-to-image person retrieval, which is a training-free framework could be directly applied to different text-to-image retrieval tasks.}
\label{fig:main}
\end{figure*}

\section{Related Works}

\subsection{Text-to-Image Person Retrieval}

Text-to-image person retrieval is part of person re-identification
\cite{zhao2025mos,liu2025try,zhao2025ccup,xu2025identity,xu2026bit,xu2026cmcc,ma2025omniperson}, aims to retrieve pedestrian images using textual descriptions, which is important for applications like surveillance and social media analysis \cite{CUHK-PEDES, ding2021semantically, bai2023text, cao2024empirical}. Early works aligned global visual-textual features in a shared space \cite{zheng2020dual, zhang2018deep,liu2025looking}, with instance loss \cite{zheng2020dual} modeling intra-modal distribution and TIMAM \cite{sarafianos2019adversarial} learning modality-invariant features.

To handle fine-grained matching challenges from high intra-class variations, part-level alignment methods were proposed. These include explicit part detection and alignment \cite{ding2021semantically, wang2020vitaa, li2023dcel,yuan2025modality,li2025text}, and implicit noun-body correspondence learning \cite{shao2023unified}. Some works \cite{wang2020high, chen2022tipcb} segment images and adaptively align text. Semantically aligned networks \cite{ding2021semantically} and multi-level matching frameworks like TIPCB \cite{chen2022tipcb} improve alignment by addressing modality gaps. IVT \cite{shu2022see} uses a unified network and semantic alignment strategies; IRRA \cite{IRRA} enhances global matching via local relation reasoning. RDE \cite{RDE} focuses on hard negatives and noisy correspondence, while DSSL \cite{zhu2021dssl} separates person and surroundings. CADA \cite{lin2024cross} builds bidirectional associations by addressing image-text discrepancies. RaSa \cite{bai2023rasa} enhances robustness with relation and sensitivity-aware representation learning.

\subsection{LLM-Driven Person Re-Identification}

Although prior works have focused on multimodal alignment, many suffer from poor generalization. To address this, Large Language Models (LLMs) have been introduced to improve person Re-ID via data generation and feature enhancement.

For data augmentation, LLMs are used to generate synthetic texts or image-text pairs \cite{yang2024mllmreid}. LV-ReID \cite{xia2025lv} combines vision-language alignment via BLIP-2 \cite{li2022blip} with a Q-Former to bridge modalities. TVI-LFM \cite{hu2024empowering} extracts and aligns features from LLM-generated texts to reduce domain gaps. TBPS-CLIP \cite{cao2024empirical} applies basic training tricks to CLIP for lightweight performance. LVLM-ReID \cite{wang2024large} uses LLM prompts to generate semantic tokens capturing pedestrian appearance. \cite{10879282} proposes a human-LLM collaborative method to inject textual descriptions into Re-ID datasets.

Beyond generation, LLMs are also used to enhance feature representation through semantic and cross-modal reasoning \cite{HAM, zuo2024plip, fu2022large}. SPAC \cite{zuo2024plip} combines ResNet101-FPN \cite{he2016deep, lin2017feature} and GPT2 \cite{radford2019language} to generate rich captions. HAM \cite{HAM} enables LLMs to learn human-style annotations for large-scale auto-labeling. CLIP \cite{radford2021learning} is used for attribute recognition with heuristic templates \cite{shao2023unified, HAM}. In \cite{yang2023towards}, diffusion models \cite{rombach2022high} synthesize images, and BLIP-2 \cite{li2022blip} generates captions. LUPerson-MLLM \cite{tan2024harnessing} builds captions using LLM-driven multi-turn dialogues, while LUPerson-T \cite{shao2023unified} generates pseudo-descriptions with a divide-conquer-combine CLIP strategy.

\section{Methods}

\subsection{Expression Drift and Semantic Invariance}
In text-to-image retrieval tasks, natural language serves as a rich medium of semantic description. However, due to the high flexibility and diversity of language, the same visual concept can often be described in various textual forms. For instance, ``a man wearing a blue jacket" and ``a man with a blue jacket" are semantically equivalent, yet due to differences in vocabulary and syntax, some words and order changes may cause the features obtained by the model to undergo feature shift, and the overall features can be affected by some irrelevant words, which impairs the performance of cross-modal matching, as shown in Fig.\ref{fig:intro}.

To better understand the feature shift caused by such variations in expression, we formulate the problem as follows. Given an original text description \( T \) with its corresponding embedding \( \mathbf{v}_T \), and a reformulated version \( T' \) with embedding \( \mathbf{v}_{T'} \), although \( T \approx T' \) semantically, we often observe:
\begin{equation}
\|\mathbf{v}_T - \mathbf{v}_{T'}\|_2 \gg 0 
\end{equation}

Furthermore, due to expression differences, the reformulated text may focus on different semantic cues, leading to shifts in the attention of the encoded features. As the first two lines in Fig.\ref{fig:intro}, the original description \( T \) may emphasize the appearance of the lower body (e.g., pants), while the reformulated version \( T' \) may highlight the upper body (e.g., coat), resulting in the following observations:
\begin{equation}
\mathbf{v}_T \cdot \mathbf{e}(\text{pant}) > \mathbf{v}_{T'} \cdot \mathbf{e}(\text{pant}), \quad
\mathbf{v}_{T'} \cdot \mathbf{e}(\text{coat}) > \mathbf{v}_{T} \cdot \mathbf{e}(\text{coat})
\end{equation}
where $v$ is the sentence vector, $e$ is the word embedding.

That is, despite referring to the same visual content, there exists a significant distance between \( T \) and \( T' \) in the feature space. We define this phenomenon as \textbf{Expression Drift}, which introduces instability and inconsistency in image-text alignment tasks.

On the other hand, the prior study~\cite{hu2024empowering,yuan2025poses,yuan2025neighbor} have demonstrated that analogical reasoning in embedding spaces can be represented through vector arithmetic. For example:
\begin{equation}
\mathbf{e}(\text{father}) - \mathbf{e}(\text{male}) + \mathbf{e}(\text{female}) \approx \mathbf{e}(\text{mother})
\end{equation}

Similarly, in the multi-modal embedding space, this property can be observed as:
\begin{equation}
\mathbf{v}_{\text{img}}(\text{black coat}) - \mathbf{v}(\text{black}) + \mathbf{v}(\text{red}) \approx \mathbf{v}_{\text{img}}(\text{red coat})
\end{equation}

These properties inspire us to introduce a compensation mechanism that captures \textbf{Semantic Invariance} — ensuring that different textual expressions conveying the same semantics are aligned consistently in the embedding space.

\subsection{LLM Collaborative Multi-View Reformulation} \label{reformulation}

To alleviate the inconsistency caused by \textbf{Expression Drift}, we propose a LLM Collaborative \textbf{Multi-View Reformulation (MVR)} mechanism. The core idea is to leverage the generative capability of large language models (LLMs) like Grok\cite{grok3}, ChatGPT\cite{openai2024gpt4o}, Qwen\cite{Qwen2.5-VL}, DeepSeek\cite{guo2025deepseek} etc.,to introduce \textbf{linguistic diversity} while preserving the core semantics, thus enhancing the robustness of textual features against natural language variation.

It is designed with two strategies to enhance the stability of textual representations in the embedding space:

\noindent\textbf{Key Feature Consistency}: The reformulated descriptions are constrained to retain visually key semantic words (e.g., colors, body parts, gender) to ensure semantic stability.

\noindent\textbf{Controlled Diversity Imagination}: Within the bounds of real-world plausibility, LLMs are guided to generate diverse expressions from different perspectives and styles, enhancing the generalization of the model to unseen phrasings.

\subsubsection{Strategy A: Key Feature Consistent Reformulation} 

\hfill \break
This strategy explicitly instructs the language model to \textbf{preserve a list of visually significant keywords} from the original description. 

To get the keywords we proposed \textbf{CLIP-based Key Words Extraction.}
Given a textual input \( T = \{w_1, w_2, \dots, w_n\} \), we aim to extract a set of key semantic tokens \(\mathcal{K}_T\) that are most relevant to the overall sentence-level embedding. We compute the importance of each token by measuring the cosine similarity between its embedding and the global sentence embedding using CLIP~\cite{radford2021learning}. Specifically, we define:
\begin{equation}
\mathcal{K}_T = \{ w_i \in T \mid \text{CosSim}(\mathbf{e}_i, \mathbf{v}_T) \geq \delta \}
\end{equation}
where, \(\mathbf{v}_T \in \mathbb{R}^d\) denotes the global sentence-level embedding obtained from CLIP's text encoder, \(\mathbf{e}_i \in \mathbb{R}^d\) is the embedding of word \(w_i\), extracted from the token embedding layer. The cosine similarity is defined as:
\begin{equation}
\mathcal{K}_T = \{ w_i \mid \tilde{\alpha}_i \geq \delta \}
\end{equation}

$\delta$ is the threshold to help retain key words. This is just a \textbf{simple and fast} conduction to get keywords, since our focus is to \textbf{validate the idea}, \textbf{NOT} the \textbf{keyword extraction accuracy}. If use a better keywords extractors could be better, but it is just an extra gain, which not affect the conclusion. This choice is empirical and may vary with the dataset and task.

With these keywords, when generating the reformulated description \( T'_a \), the model must satisfy:
\begin{equation}
T^a = \text{LLM}(T \mid \text{P}_{\text{key}}; \forall k \in \mathcal{K}_T, \; k \in T^a)  
\end{equation}

where, $\text{P}_{\text{key}}$ is the prompt with keywords. That is, the reformulated sentence must explicitly retain all keywords in \( \mathcal{K}_T \), ensuring the semantic anchor is preserved and minimizing potential feature drift.

\subsubsection{Strategy B: Diversity-aware Reformulation}

\hfill \break
This strategy allows the LLM to \textbf{generate reformulations with free-form expressions under realistic constraints}, enabling diverse perspectives to emerge without semantic distortion. We guide the language model with:
\begin{equation}
T^b = \text{LLM}(T \mid \text{P}_{\text{diverse}}), \quad \text{s.t. } \text{D}(T, T^b) \leq \tau
\end{equation}
where, $\text{P}_{\text{diverse}}$ is the prompt with diversity, \(\text{D}(T, T^b)\) denotes the semantic diversity between the original and reformulated texts, denote as temperature \(\tau\), which is the LLM's temperature, controls semantic faithfulness. This constraint ensures reformulated sentence remains anchored to the original text.

\begin{table*}[tp]
\caption{Comparisons with state-of-the-art ReID methods under the traditional evaluation setting. \textbf{Bold} indicates performance improvement over the corresponding baseline. \textbf{Gray highlight denotes the state-of-the-art result on that dataset}.}
\vspace{-0.8em}
\label{tab:sota-traditional}
\centering
\resizebox{0.975\linewidth}{!}{
\begin{tabular}{c|c|cccc|cccc|cccc}
\hline
\multirow{2}{*}{Method} & \multirow{2}{*}{Venue} & \multicolumn{4}{c|}{RSTPReid} & \multicolumn{4}{c|}{CUHK-PEDES} & \multicolumn{4}{c}{ICFG-PEDES}  \\ 
\cline{3-14} 
 &  & R1 & R5 & R10 & mAP & R1 & R5 & R10 & mAP & R1 & R5 & R10 & mAP \\ 
\hline
CAIBC \cite{wang2022caibc} & ACM MM '22 &47.35 &69.55 &79.00& - & 64.43 & 82.87 & 88.37 & - &- &- &- &-  \\
LGUR \cite{shao2022learning} & ACM MM '22 & 47.95 & 71.85 & 80.25 & -  & 65.25 & 83.12 & 89.00 & - & 59.02 & 75.32 & 81.56 & - \\
DCEL \cite{li2023dcel} & ACM MM '23 & 61.35 & 83.95 & 93.46 & - & 73.71 & 89.33 & 93.46 & - & 64.88 & 81.34 & 86.72 & -  \\
LCR$^2$S \cite{yan2023learning} & ACM MM '23 & 54.95 & 76.65 & 84.70 & 40.92 & 67.36 & 84.19 & 89.62 & 59.20 & 57.93 & 76.08 & 82.40 & 38.21  \\

APTM \cite{yang2023towards} & ACM MM '23 & 67.50 & 85.70 & 91.45 & 52.56 & 76.53 & 90.04 & 94.15 & 66.91 & 68.51 & 82.99 & 87.56 & 41.22  \\
RaSa \cite{bai2023rasa}& IJCAI '23 & 66.90 & 86.50 & 91.35 & 52.31 & 76.51 & 90.29 & 94.25 & 69.38 & 65.28 & 80.40 & 85.12 & 41.29  \\

CFine \cite{yan2023clip} & TIP '23 & 50.55 &72.50 &81.60 & - & 69.57 & 85.93 & 91.15 & - & 60.83 & 76.55 & 82.42 & -  \\ 
VGSG \cite{he2023vgsg} & TIP '23 & - & - & - & - & 67.52 & 84.37 & 90.26 & - &60.64 & 76.01 & 82.01  \\
UniPT \cite{shao2023unified} & ICCV '23 & 51.85 & 74.85 & 82.85 & - & 68.50 & 84.67 & 90.38 & - & 60.09 & 76.19 & 82.46 & -  \\
CADA \cite{lin2024cross} & TMM '24 & 67.70 & 84.60 & 89.75 & 49.95 & 77.20 & 90.68 & 93.92 & 68.45 & 67.38 & 81.34 & 85.64 & 37.81  \\ 
TBPS-CLIP \cite{cao2024empirical}  & AAAI '24 & 61.95 & 83.55 & 88.75 & 48.26 & 73.54 & 88.19 & 92.35 & 65.38 & 65.05 & 80.34 & 85.47 & 39.83  \\
CFAM \cite{zuo2024ufinebench} &  CVPR '24 & 60.51 & 82.85 & 89.71 & 47.64 & 73.67 & 89.71 & 93.57 & 65.94 & 63.57 & 80.57 & 86.32 & 38.34  \\
UMSA  \cite{zhao2024unifying} & AAAI '24 & 63.40 & 83.30 & 90.30 & 49.28 & 74.25 & 89.83 & 93.58 & 66.15 & 65.62 & 80.54 & 85.83 & 38.78  \\
LSPM \cite{li2024learning} & TMM '24 & - & - & - & -  & 74.38 & 89.51 & 93.42 & 67.74 & 64.40 & 79.96 & 85.41 & 42.60 \\
IRLT \cite{liu2024causality} & AAAI '24 & 61.49 & 82.26 & 89.23 & -  & 74.46 & 90.19 & 94.01 & - & 64.72 & 81.35 & 86.31 & - \\
FSRL \cite{wang2024fine} & ICMR '24 & 60.20 & 81.40 & 88.60 & 47.38 & 74.65 & 89.77 & 94.03 & 67.49 & 64.01 & 80.42 & 85.56 & 39.64  \\
Propot \cite{yan2024prototypical} & ACM MM '24 & 61.87 & 83.63 & 89.70 & 47.82 & 74.89 & 89.90 & 94.17 & 67.12 & 65.12 & 81.57 & 86.97 & 42.93  \\

PLOT \cite{park2024plot} & ECCV '24 & 61.80 & 82.85 & 89.45 & - & 75.28 & 90.42 & 94.12 & - & 65.76 & 81.39 & 86.73 & -  \\
AUL \cite{li2024adaptive} & AAAI '24 & 71.65 & 87.55 & 92.05 & -  & 77.23 & 90.43 & 94.25 & - & 69.16 & 83.32 & 88.37 & - \\
F-WoRA \cite{sun2025data} & WWW '25 & 66.85 & 86.45 & 91.10 & 52.49 & 76.38 & 89.72 & 93.49 & 67.22 & 68.35 & 83.10 & 87.53 & 42.60  \\
\hline
IRRA \cite{IRRA} &  CVPR '23 & 60.20 & 81.30 & 88.20 & 47.17  & 73.38 & \textbf{89.93} & 93.71 & 66.10 & 63.46 & 80.25 & 85.82 & 38.06 \\
\rowcolor[HTML]{EAFAF1} 
IRRA + \textbf{ours} & - & \textbf{63.10}  & \textbf{82.40}  & \textbf{89.40}  & \textbf{49.21} & \textbf{74.35}  & 89.56 & \textbf{93.83}  & \textbf{66.98} & \textbf{64.51} & \textbf{80.40} & \textbf{85.83} & \textbf{38.70}  \\
\hline
RDE \cite{RDE} & CVPR '24 & 65.35 & 83.95 & 89.90 & 50.88  & 75.94 & 90.14 & 94.12 & 67.56 & 67.68 & 82.47 & 87.36 & 40.06 \\
\rowcolor[HTML]{EAFAF1} 
RDE + \textbf{ours} & -  & \textbf{67.60} & \textbf{85.15} & \textbf{90.10} & \textbf{52.07} & \textbf{76.23} & \textbf{90.27} & \textbf{94.36} & \textbf{68.10} & \textbf{68.11} & \textbf{82.48} & \textbf{87.40} & \textbf{40.60}  \\
\hline
HAM(IRRA) \cite{HAM} & CVPR '25 & 70.80 & 87.50 & 92.40 & 54.52  & 76.25 & 90.63 & \textbf{94.74} & 68.58 & 68.25 & 82.88 & 87.69 & 41.84 \\
\rowcolor[HTML]{EAFAF1} 
HAM(IRRA) + \textbf{ours} & - & \textbf{71.05} & \textbf{87.85} & \cellcolor{gray!25}{\textbf{92.70}} & \textbf{55.15} & \textbf{76.79}  & \textbf{90.92} & 94.23 & \textbf{68.99} & \textbf{68.60} & \textbf{83.17} & \textbf{87.78} & \textbf{42.58}  \\
\hline
HAM(RDE) \cite{HAM} & CVPR '25 & 71.50 & 86.50 & 91.15 & 55.34  & 77.99 & 91.34 & 95.03 & 69.72 & 69.95 & 83.88 & 88.39 & 42.72 \\ 
\rowcolor[HTML]{EAFAF1} 
HAM(RDE) + \textbf{ours} & - & \cellcolor{gray!25}{\textbf{73.75}} & \cellcolor{gray!25}{\textbf{87.90}} & \textbf{92.55} & \cellcolor{gray!25}{\textbf{55.74}} & \cellcolor{gray!25}{\textbf{78.54}} & \cellcolor{gray!25}{\textbf{91.39}} & \cellcolor{gray!25}{\textbf{95.08}} & \cellcolor{gray!25}{\textbf{70.22}} & \cellcolor{gray!25}{\textbf{70.52}} & \cellcolor{gray!25}{\textbf{84.20}} & \cellcolor{gray!25}{\textbf{88.67}} & \cellcolor{gray!25}{\textbf{43.35}}   \\

\hline
\end{tabular}
}
\end{table*}

\subsection{Textual Feature Robustness Compensation} \label{3.3}

Despite the strong expressive power of current language models, textual features derived from a single query often fail to fully capture the inherent semantic richness of natural language, primarily due to \textbf{Expression Drift}---the phenomenon where semantically equivalent queries exhibit distributional discrepancies in the feature space. To mitigate this, we propose a simple yet surprisingly effective \textbf{training-free} mechanism that enhances textual representations via multi-view semantic aggregation, all while keeping the text encoder entirely frozen.

\noindent\textbf{Multi-View Textual Feature Extraction.}
Given an original query text \( T \), we employ the collaborative LLM-based mechanism proposed in Sec.~\ref{reformulation} to construct set of semantically diverse reformulations: 
\(\mathcal{R}^{\text{text}} = \{T^{a}_{i}, T^{b}_{i}\}_{i=1}^{N}\), where $N$ is the number of reformulations.

Each textual reformulation, together with the original input, is encoded via a frozen text encoder \( \boldsymbol{\Phi}(\cdot) \)\cite{hu2024empowering}, and the obtained multi-view diversity textual features are aggregated together to get robust compensation textual feature, as formulated:
\begin{equation}
    \hat{\mathbf{v}}^{\text{text}} = \mathbf{\Phi}(T) + \alpha \cdot \textbf{Mean}(\mathbf{\Phi}(R_i)), \forall R_i \in \mathcal{R}^{\text{text}}
\label{eq:alpha}
\end{equation}
where $\alpha$ is a hyperparameter controlling the impact of multi-view semantic for text feature.

This feature compensation preserves the identity-preserving semantics of original query while injecting auxiliary perspectives that improve the robustness and expressiveness of the feature in downstream retrieval tasks.

Moreover, this formulation requires no additional training, gradients, or model updates. Instead, it operates entirely in latent feature space, distilling multi-view semantics into a unified representation. Despite its simplicity, the mechanism significantly improves retrieval robustness, especially under paraphrased, noisy, or domain-shifted queries. We attribute this to its ability to capture what we term \textbf{semantic echoes}---the recurrent core meaning reverberating across diverse linguistic expressions.

\subsection{High-level Semantic Compensation}
While vision encoders are adept at capturing low-level structures and appearance patterns, they often lack sensitivity to high-level abstract semantics, which are crucial for fine-grained cross-modal alignment. Therefore, we propose a semantic compensation mechanism that projects visual signals into the textual domain via vision-language modeling and reinforces them through multi-view language augmentation.

\noindent\textbf{Visual Description Extraction.}
Given an input image \( I \), we first generate a high-level semantic description using a pre-trained vision-language model (VLM):
\begin{equation}
    S = \text{VLM}(I)
\end{equation}

where \( S \) is a textual summary of the image content, capturing not only object presence and spatial configurations but also contextual cues and abstract meanings. This serves as a semantic compensation for the image in language space.

\noindent\textbf{Multi-View Reformulation.}
To comprehensively capture the intrinsic semantic variability of \( S \), we use the same way discussed in (Sec.~\ref{3.3}) to generate a set of diverse yet semantically aligned textual variants \(\mathcal{R}^{\text{img}} = \{S^{a}_{i}, S^{b}_{i}\}_{i=1}^{N}\).

Then, with a frozen vision encoder \( \boldsymbol{\Psi}(\cdot) \) and text encoder \( \boldsymbol{\Phi}(\cdot) \), we can get high-level semantic compensation feature.
\begin{equation}
    \hat{\mathbf{v}}^{\text{img}} = \mathbf{\Psi}(I) + \beta \cdot \textbf{Mean}(\mathbf{\Phi}(S_i)), \forall S_i \in \mathcal{R}^{\text{img}}
\label{eq:beta}
\end{equation}
where $\beta$ is a hyperparameter controlling the contribution of multi-view semantic for image feature.

This cross-modal compensation mechanism enables the visual representation to benefit from the compositional power of language, effectively aligning the visual semantics with textual space.

\section{Experiments}
\subsection{Datasets and Metrics}
\textbf{CUHK-PEDES}\cite{CUHK-PEDES} serves as a foundational benchmark for research in the text-to-image person retrieval. It comprises 40,206 pedestrian images corresponding to 13,003 distinct identities, with each image accompanied by two independently annotated textual descriptions. Following the official data split, the training set includes 34,054 images and 68,108 textual descriptions covering 11,003 identities. The testing set involves 3,074 images and 6,156 descriptions from 1,000 identities, while a separate validation set consists of 3,078 images from another 1,000 identities.

\noindent\textbf{ICFG-PEDES}\cite{ding2021semantically} is a large-scale text-to-image person retrieval dataset, consisting of  54,522 pedestrian images associated with 4,102 distinct identities. Each image is annotated with a single fine-grained textual description, making it more focused and discriminative than CUHK-PEDES. The dataset is split into a training set with 34,674 image-text pairs from 3,102 identities and a testing set with 19,848 pairs corresponding to the remaining 1,000 identities.

\noindent\textbf{RSTPReid}\cite{zhu2021dssl} is a text-to-image person retrieval dataset designed to reflect real-world surveillance scenarios. It comprises 20,505 pedestrian images collected from 15 distinct cameras, covering 4,101 unique identities. Each identity is represented by five images captured from different viewpoints, and every image is annotated with two textual descriptions, resulting in a total of 41,010 image-text pairs. Following the official dataset protocol, the training set includes 18,505 images and 37,010 descriptions from 3,701 identities. Both the validation and testing sets are composed of 1,000 images and 2,000 descriptions, each corresponding to 200 identities.

\begin{table}[tp]
\caption{Ablation study of our proposed query and gallery feature compensation on the RSTPReid dataset with IRRA baseline.}
\vspace{-0.8em}
\label{tab: ablation}
\centering
\resizebox{0.975\linewidth}{!}{
\begin{tabular}{cc|cc|cccc}
\hline
\multicolumn{2}{c|}{Query} & \multicolumn{2}{c|}{Gallery}  & \multicolumn{4}{c}{RSTPReid} \\ 
$P_{key}$& $P_{diverse}$ & $P_{key}$ &  $P_{diverse}$ & R1 & R5 & R10 & mAP \\
\hline
\ding{55} & \ding{55}& \ding{55} & \ding{55} & 60.20 & 81.30 & 88.20 & 47.17 \\
\ding{52} & \ding{55} & \ding{52} & \ding{55} & 61.80 & 82.95 & 89.30 & 48.40 \\
\ding{55} & \ding{52} & \ding{55} & \ding{52} & 61.90 & 82.00 & 89.50 & 48.58 \\
\ding{55} &\ding{55} & \ding{52} & \ding{52} & 60.95 & 82.95 & 89.25 & 47.95 \\
\ding{52}& \ding{52}& \ding{55} & \ding{55} & 61.35 & 81.75 & 89.15 & 48.43 \\
\rowcolor[HTML]{EAFAF1} 
 \ding{52} &\ding{52}  & \ding{52} & \ding{52} & 63.10 & 82.40 & 89.40 & 49.20 \\
 \hline
\end{tabular}
}
\end{table}

\textbf{Evaluate Metrics.} In line with prior studies \cite{RDE, IRRA}, we employ Rank-k accuracy (with k = 1, 5, 10) and mean Average Precision (mAP) as the primary evaluation metrics across all three datasets. These metrics collectively assess the retrieval accuracy and overall matching quality. A higher value in either Rank-k or mAP indicates superior performance.

\subsection{Implementation Details}
\begin{figure}
\centering
\includegraphics[width=\linewidth]{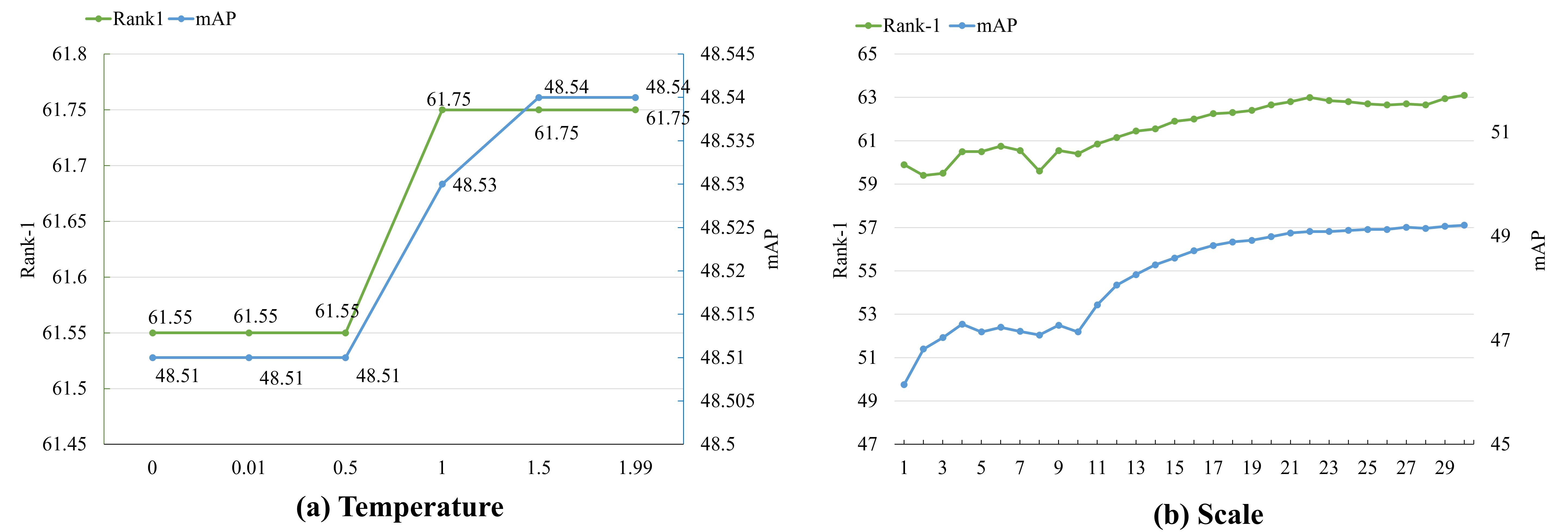}
\caption{(a)Impact of different temperature $\tau$ on RSTPReid dataset with IRRA. (b)Impact of query feature compensation scale on RSTPReid dataset with IRRA baseline.}
\label{fig:tem_scale}
\end{figure}

All experiments were conducted on one NVIDIA RTX 4090 GPU and one Intel I9-13900K CPU. For LLM collaborative reformulation, we employed TVI-LFM \cite{hu2024empowering} to generate the caption of the image gallery and Qwen2.5-VL-32B \cite{Qwen2.5-VL} to reformulate query and gallery captions, with the decoding temperature set to 0.01. Specifically, we designed two distinct prompt templates to guide the reformulation process for both queries and image-derived captions. Each template yields 15 reformulated variants (just in one QA), resulting in a total of 30 semantic reformulations per input instance (it can also use less reformulations as show in \ref{fig:tem_scale} (b) for faster speed) 
For empirical comparisons, we adopted four representative baseline methods: IRRA~\cite{IRRA}, RDE~\cite{RDE}, HAM(IRRA)~\cite{HAM}, and HAM(RDE), evaluating each using the default configurations and pretrained weights provided in their official open-source implementations.
The threshold $\delta$ with default -0.03, and it helps retain some neutral words, avoiding overly strict filtering. It is the average of GPT labeled key words of all datasets. The hyperparameter $\alpha$ in Eq.~\ref{eq:alpha} is set to 0.75, and $\beta$ in Eq.~\ref{eq:beta} is set to 0.3, both selected via grid search on the validation set.

\begin{figure}
\centering
\includegraphics[width=\linewidth]{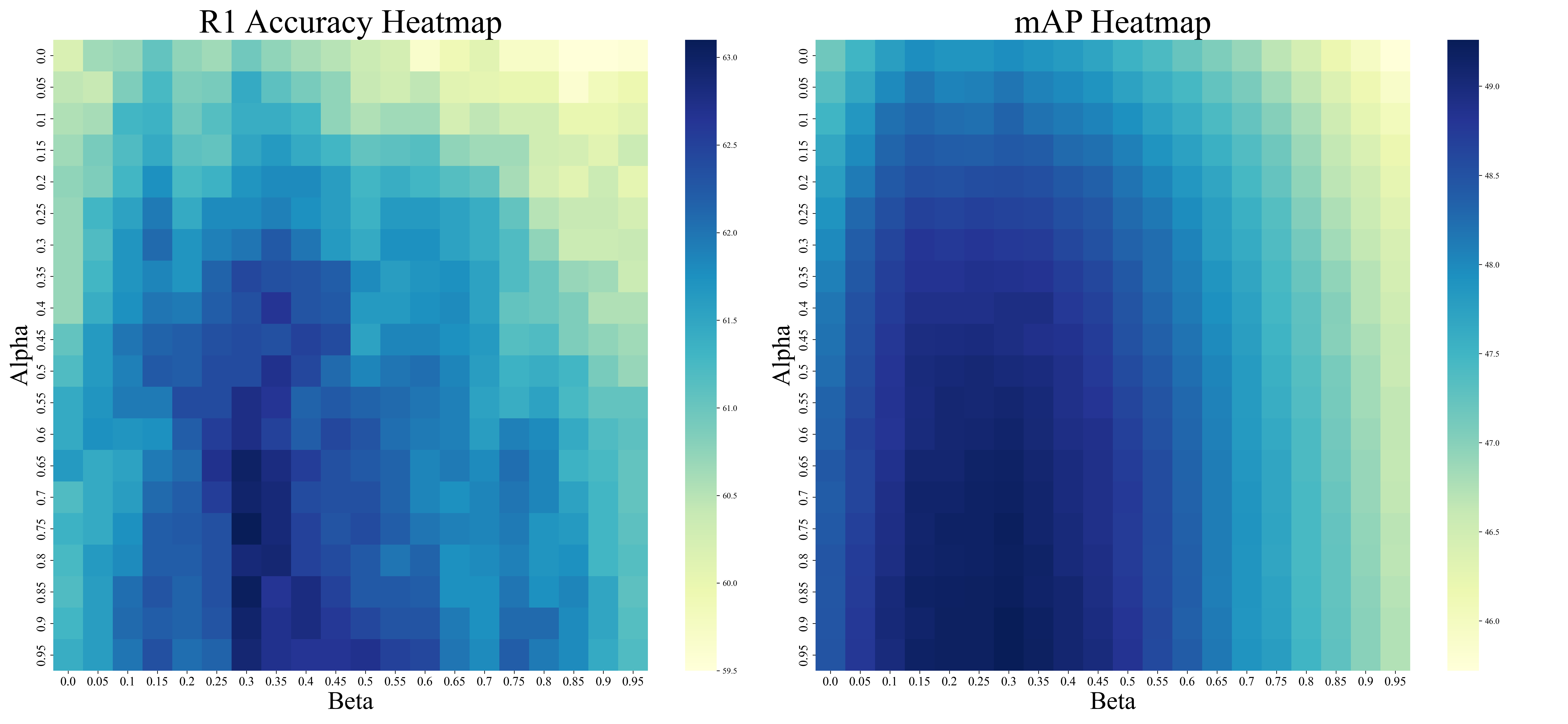}
\caption{Heatmap of R1 accuracy and mAP on the RSTPReid dataset with IRRA baseline.}
\label{fig:heatmap}
\end{figure}

% \begin{figure}
% \centering
% \includegraphics[width=0.95\linewidth]{figs/scale.png}
% \caption{Impact of query feature compensation scale on the RSTPReid dataset with IRRA baseline.}
% \label{fig:scale}
% \end{figure}
\subsection{Comparisons with State-of-the-Art Methods}

We conduct a thorough experimental evaluation comparing our method with recent state-of-the-art approaches across three standard benchmarks: RSTPReid, CUHK-PEDES, and ICFG-PEDES. The comprehensive results, presented in Table ~\ref{tab:sota-traditional}, demonstrate that our method achieves consistent and significant improvements when integrated with various baseline frameworks, establishing new \textbf{state-of-the-art} performance across multiple evaluation metrics.

\subsection{Ablation Study}

To validate the effectiveness of our proposed feature compensation strategy, we conduct a comprehensive ablation study on the RSTPReid dataset using the IRRA baseline, as shown in Table~\ref{tab: ablation}. When no compensation is applied to either the query or gallery features, the baseline achieves 60.20\% R1 accuracy and 47.17\% mAP.

Introducing compensation using keyword-based prompts ($P_{\text{key}}$) leads to performance improvements, achieving 61.80\% R1 and 48.40\% mAP. Likewise, employing diverse prompts ($P_{\text{diversity}}$) yields further gains, with R1 reaching 61.90\% and mAP increasing to 48.58\%.

When compensation with both $P_{\text{key}}$ and $P_{\text{diversity}}$ is applied solely to the query side, performance rises to 61.35\% R1 and 48.43\% mAP, indicating that enhancing the semantic richness of query features is beneficial. Similarly, applying compensation only to the gallery side results in 60.95\% R1 and 47.95\% mAP, demonstrating that gallery-side enrichment also contributes positively.

The most significant improvement is observed when both query and gallery features are jointly compensated using both prompts, achieving 63.10\% R1 and 49.20\% mAP. This represents a relative gain of +2.90\% in R1 and +2.03\% in mAP over the baseline. These results confirm the complementary nature of the two compensation strategies and underscore their effectiveness in addressing the modality gap and semantic misalignment challenges inherent in text-to-image person retrieval.

\begin{table}[tp]
\caption{Comparison of different LLMs for query and gallery feature compensation on RSTPReid with IRRA baseline. The LLMs utilized, in sequential order, are Grok-3, DeepSeek-V3, GPT-4o-Mini, and Qwen2.5-VL-32B.}
\vspace{-0.8em}
\label{tab: different llm}
\centering
\resizebox{0.975\linewidth}{!}{
\begin{tabular}{cccc|cccc}
\hline
\multicolumn{4}{c|}{LLMs}  & \multicolumn{4}{c}{RSTPReid} \\ 
\includegraphics[width=0.03\textwidth]{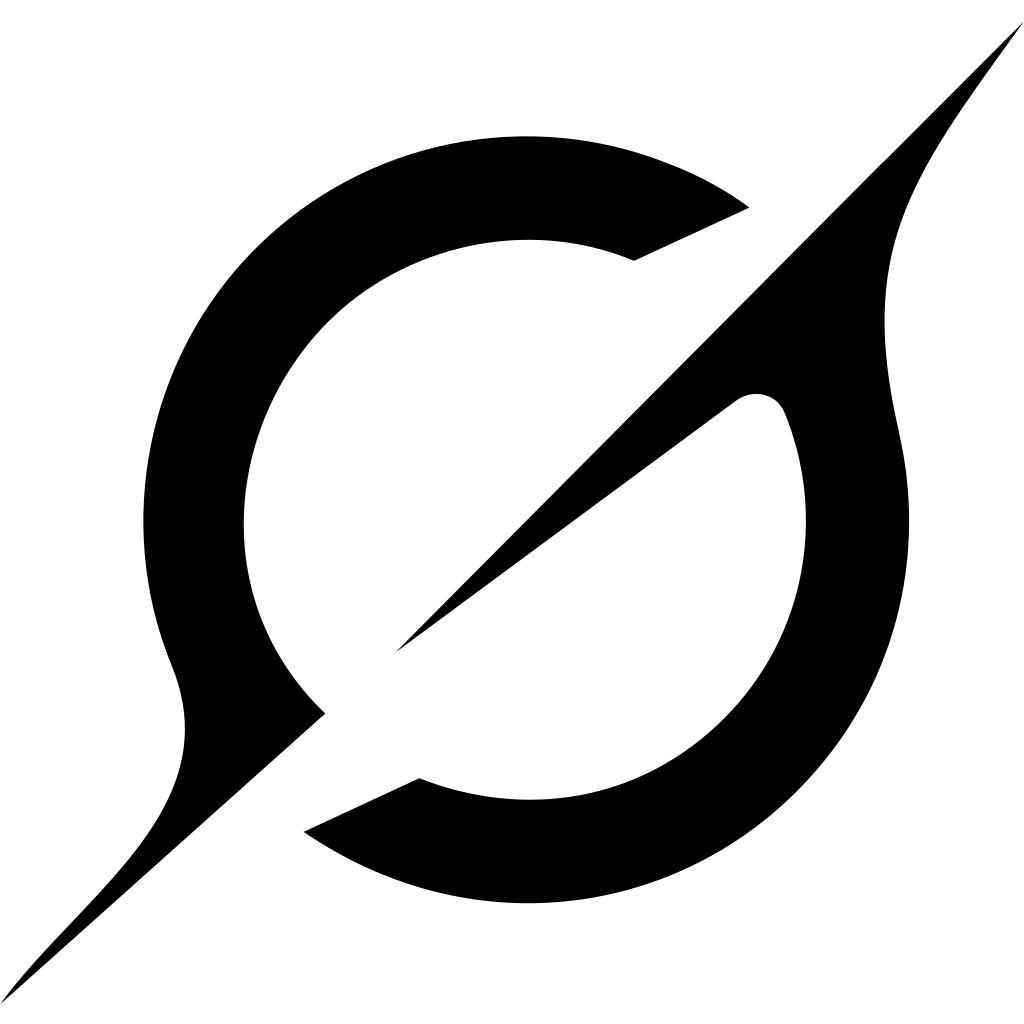} & \includegraphics[width=0.03\textwidth]{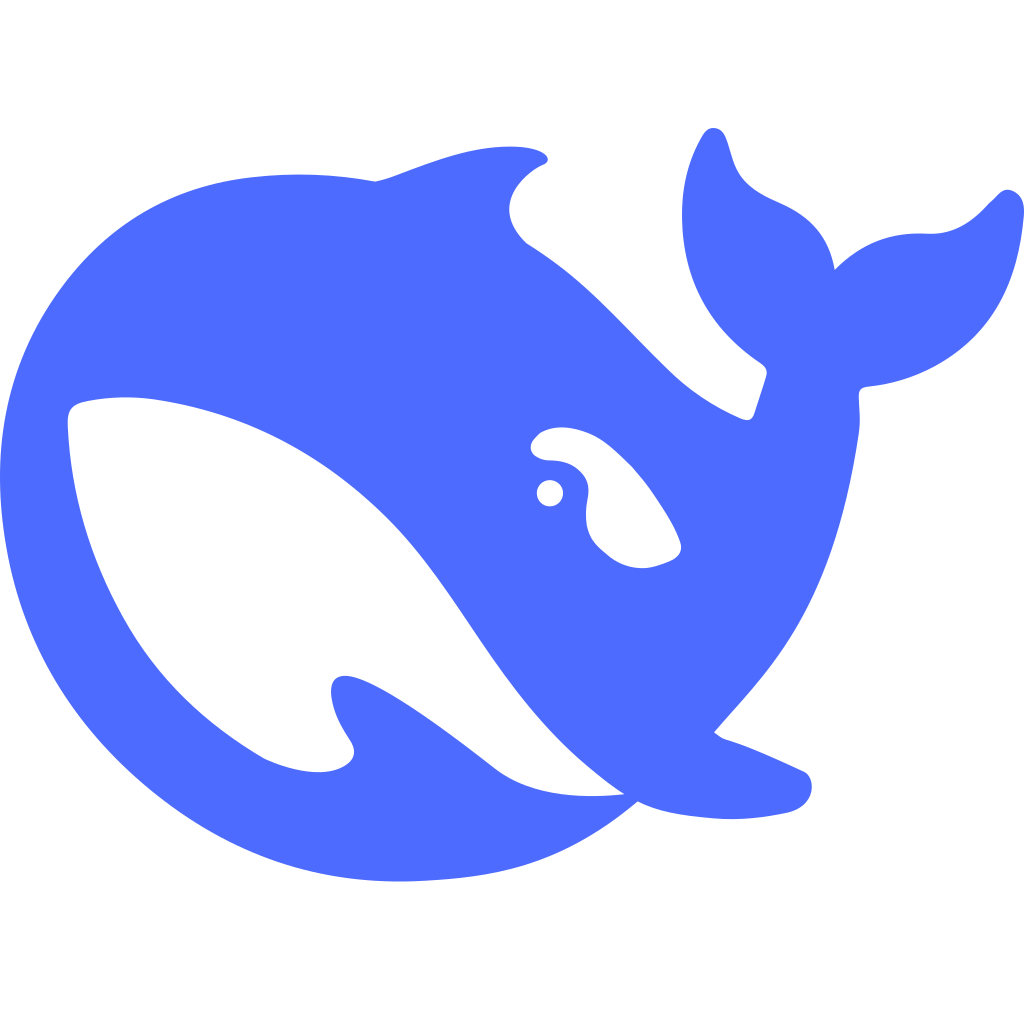} & \includegraphics[width=0.03\textwidth]{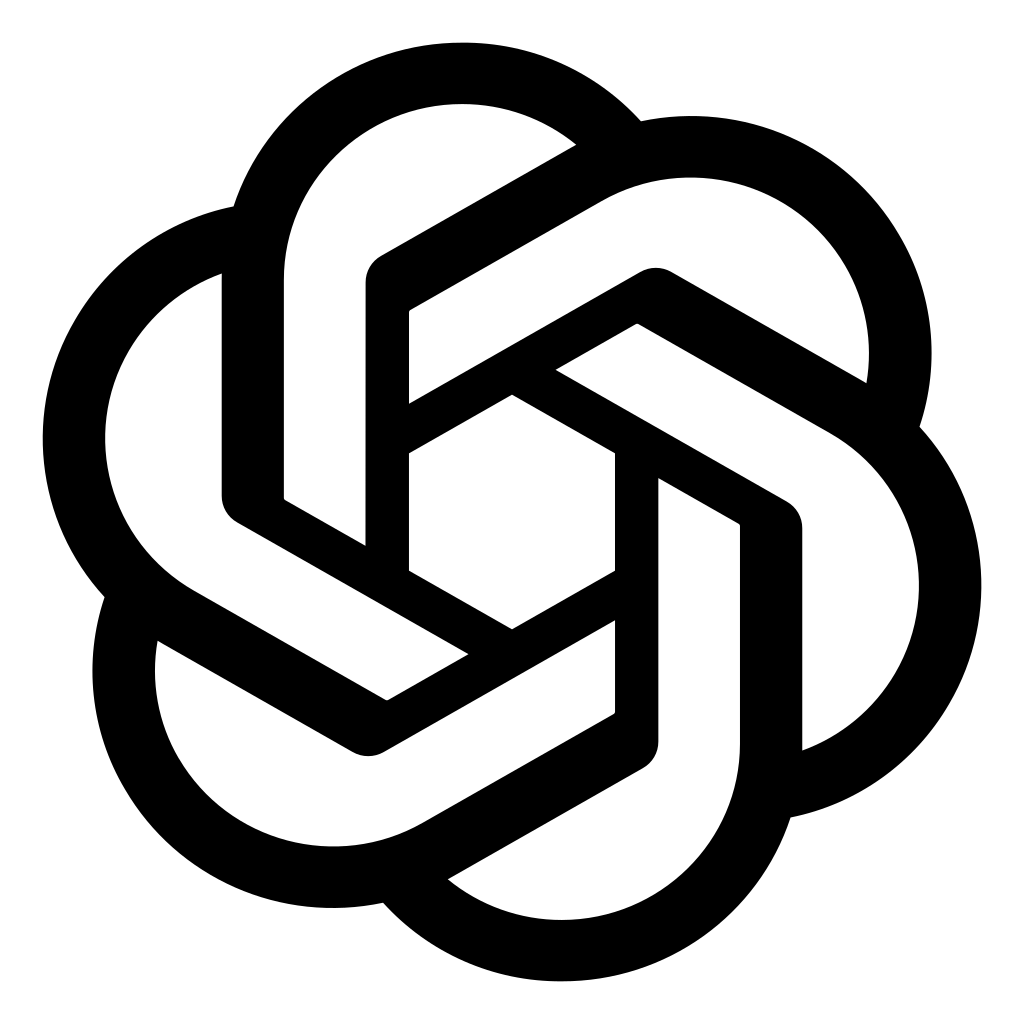}& \includegraphics[width=0.03\textwidth]{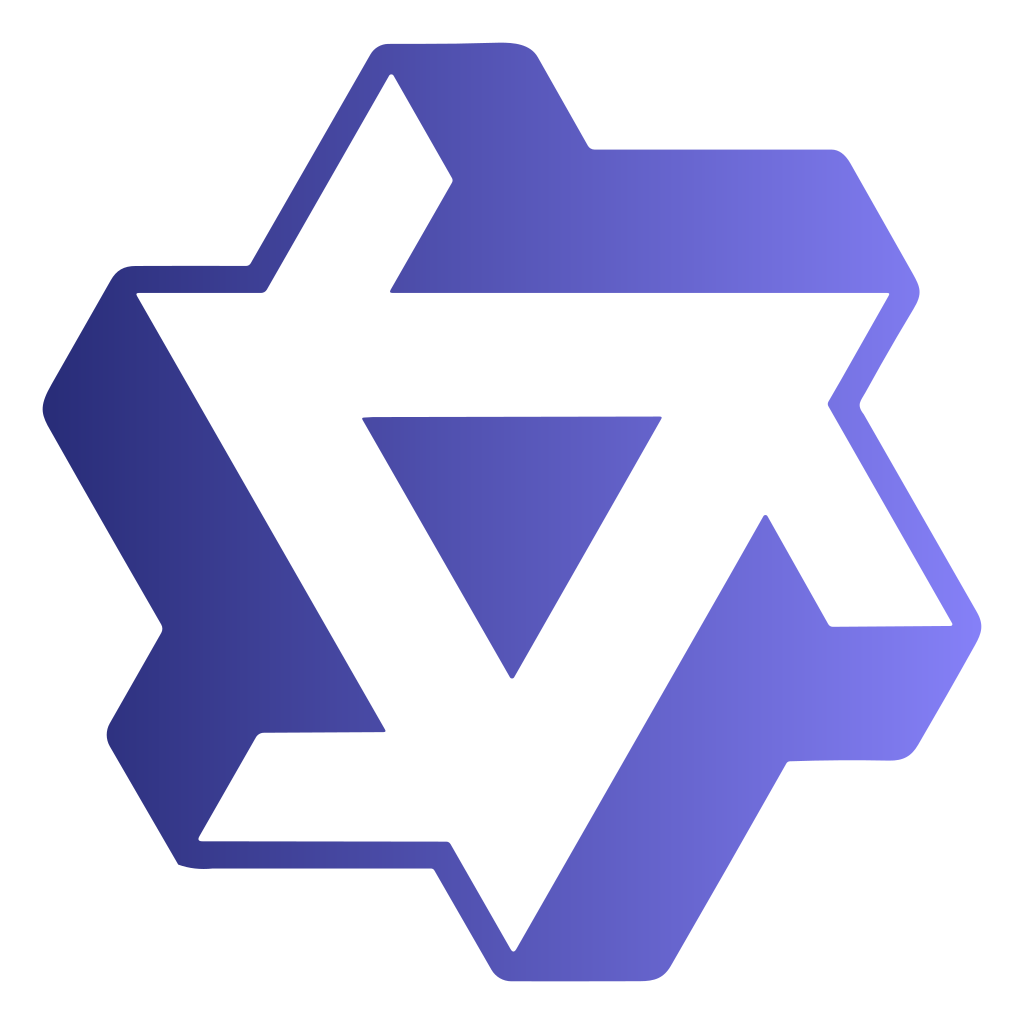}& R1 & R5 & R10 & mAP \\
\hline
\ding{55}& \ding{55}& \ding{55}& \ding{55} & 60.20 & 81.30 & 88.20 & 47.17 \\
 \hline
\ding{52} &\ding{55} &\ding{55} & \ding{55} & 60.95 & 81.15 & 88.55 & 48.08 \\
\ding{55}& \ding{52}& \ding{55}& \ding{55} & 61.10 & 82.00 & 88.95 & 48.19 \\
\ding{55}&\ding{55} & \ding{52}&\ding{55} & 60.95 & 81.85 & 89.20 & 48.18 \\
\ding{55} & \ding{55}& \ding{55}& \ding{52} & 61.55 & 81.75 & 88.9 & 48.51 \\
\hline
\ding{52} & \ding{52} & \ding{55} & \ding{55} & 61.10 & 82.50 & 89.00 & 48.47 \\
\ding{52} & \ding{52} &\ding{52} & \ding{55} & 61.45 & 81.85 & 89.05 & 48.45\\
% \ding{52} & \ding{52} & \ding{55} & \ding{55} & 61.30 & 81.85 & 89.05 & 48.29 \\
% \ding{52} & \ding{52} & \ding{52} & \ding{55} & 61.95 & 81.90 & 88.75 & 48.57 \\
% \ding{52} & \ding{52} & \ding{55} & \ding{52} & 61.30 & 82.00 & 88.95 & 48.54 \\
% \ding{52} & \ding{52} & \ding{55} & \ding{52} & 61.45 & 81.95 & 89.45 & 48.23 \\
\ding{52} & \ding{52} & \ding{52} & \ding{52} & 61.70 & 82.20 & 89.10 & 48.42\\
 \hline
\end{tabular}
}
\end{table}

\begin{figure*}
\centering
\includegraphics[width=0.9\linewidth]{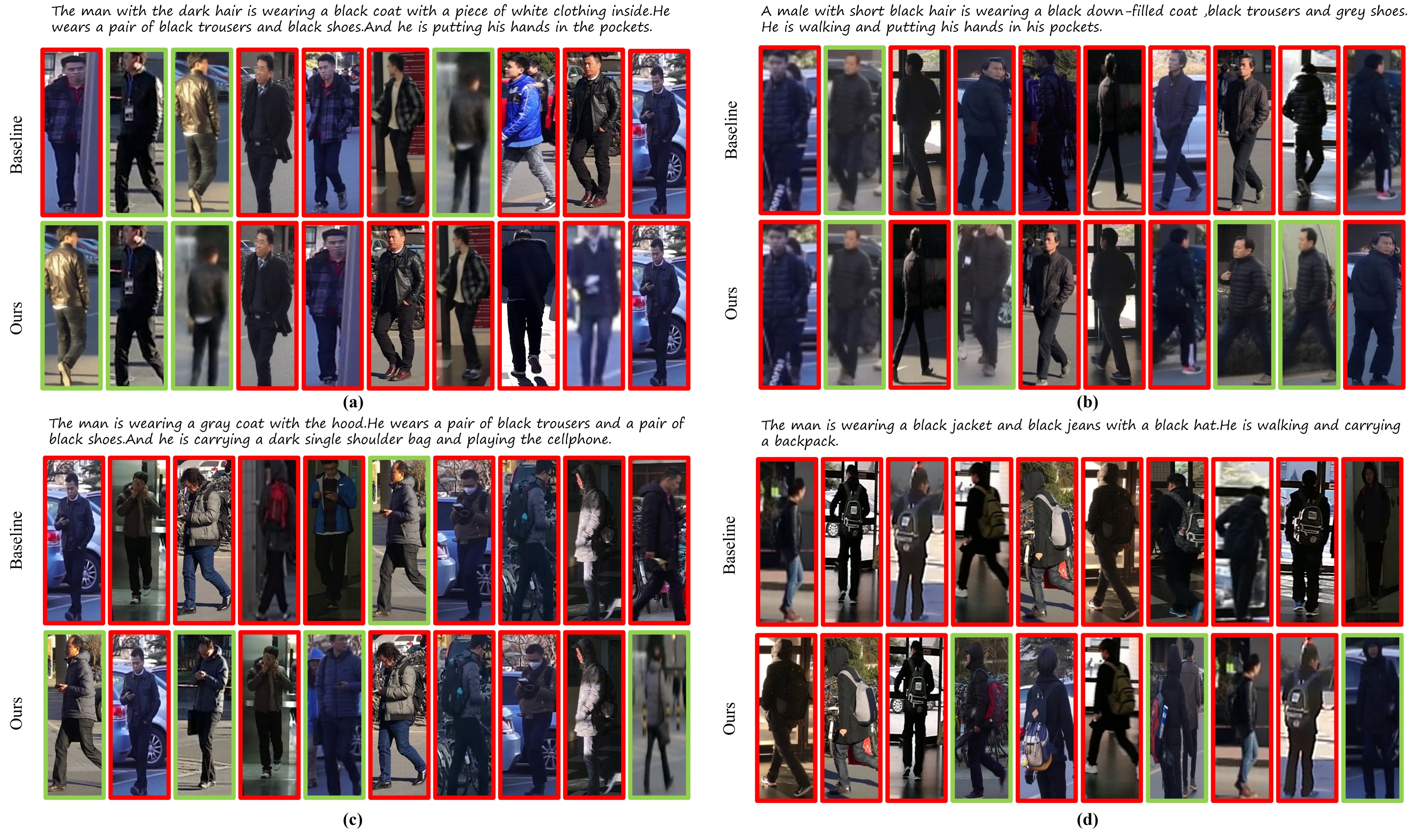}
\caption{Visual comparison of top-10 retrieved results between IRRA(baseline) and our method. Green boxes indicate correct matches, while red boxes are incorrect ones.}
\label{fig:visual}
\end{figure*}

\subsection{Hyperparameter sensitivity analysis}

\textbf{The impact of $\alpha$ and $\beta$.} Our sensitivity analysis in Fig~\ref{fig:heatmap} explores the effect of varying $\alpha$ and $\beta$ on R1 accuracy and mAP. The heatmaps show optimal performance when $\alpha \approx 0.75$ and $\beta \approx 0.3$, where R1 peaks at 63.1\% and mAP at 49.2\%. These results highlight a stable parameter range that maximizes both metrics, demonstrating the robustness of our method. Notably, the dark regions in both heatmaps align, indicating that tuning $\alpha$ and $\beta$ to these values improves retrieval accuracy and precision.

\noindent\textbf{The impact of temperature $\tau$}. To investigate the effect of temperature on caption generation, we vary the temperature of Qwen2.5-VL-32B-Instruct within the range [0,2) and evaluate its performance using the IRRA baseline. As shown in Fig.~\ref{fig:tem_scale}(a), both R1 and mAP remain stable at low temperatures, with R1 at 61.55\% and mAP at 48.51\%. These results suggest that higher temperatures introduce greater diversity in the generated captions, thereby enhancing retrieval performance.

\noindent\textbf{The scalability of our proposed MVR.} To evaluate the scalability of our method, we conducted experiments to examine the impact of varying the query feature compensation scale. The results, shown in Fig.~\ref{fig:tem_scale}(b), illustrate the performance as a function of the scale parameter. As the scale increases, both R1 accuracy and mAP consistently improve, reaching their peak values at the maximum scale of 30. R1 and mAP reach 63.1\% and 49.2\%, respectively. This upward trend demonstrates that our method benefits significantly from larger scales. Such behavior highlights the scalability of our method, as it effectively leverages higher-scale feature compensation to enhance retrieval accuracy.

\begin{figure}
\centering
\includegraphics[width=\linewidth]{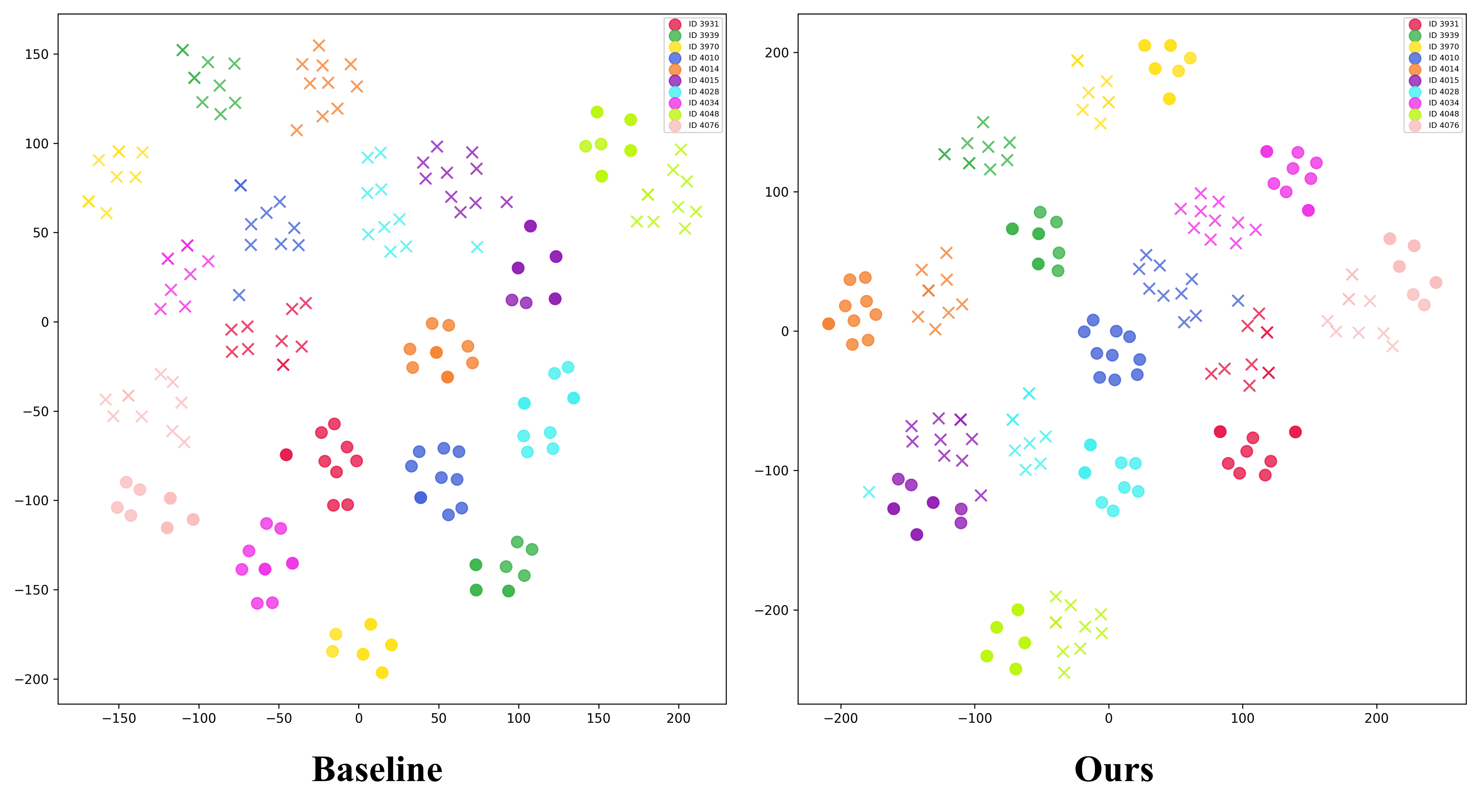}
\caption{T-sne visualizations of the baseline and ours method. The crosses represent query features and circles represent gallery features.}
\label{fig:tsne}
\end{figure}
\subsection{Effect of different LLMs}

To explore differences in LLM representations, we conducted an interesting experiment demonstrating that leveraging stylistic diversity across different LLMs enhances feature representation, validating our use of diverse rewrites.

Table~\ref{tab: different llm} presents the results of various LLMs for feature compensation on the RSTPReid dataset under the IRRA baseline. Specifically, we assess four representative LLMs: Grok-3\cite{grok3}, DeepSeek-V3 \cite{guo2025deepseek}, GPT-4o-Mini \cite{openai2024gpt4o}, and Qwen2.5-VL-32B \cite{Qwen2.5-VL}, both individually and in combination. Feature compensation is evaluated using the diversity prompt $P_{\text{diversity}}$, with each LLM producing 10 reformulated queries.

% As shown in Table~\ref{tab: different llm}, all four LLMs contribute to consistent improvements over the non-compensated baseline, verifying the effectiveness of diversity-oriented feature compensation. Among the individual models, Qwen2.5-VL-32B achieves the best performance, indicating its strong reasoning ability under the diversity prompt. Moreover, combining multiple LLMs generally leads to further gains. These findings suggest the fusion of different LLMs helps generate more robust and diverse features for text-to-image person retrieval.

As shown in Table~\ref{tab: different llm}, all four LLMs consistently improve over the baseline, confirming the effectiveness of diversity-based feature compensation. Qwen2.5-VL-32B performs best, highlighting its strong reasoning ability. Combining multiple LLMs yields further gains, suggesting enhanced robustness and diversity.

\subsection{Visualization and qualitative analysis.}

\noindent\textbf{The t-sne visualization. } 
Fig.\ref{fig:tsne} shows the distances between query and gallery. The crosses represent query features and circles represent gallery features. It showsthe great capbility of our method to align features between texts and images.

\noindent\textbf{The retrieval visualization. } 
We visualize the comparison of top-10 retrieved results between the IRRA baseline and our proposed MVR method, as illustrated in the Fig ~\ref{fig:visual}. Our method consistently retrieves more correct galleries compared to the baseline. For example, in (b), although the baseline struggles with retrieving visually similar individuals with mismatched clothing, our method accurately localizes the correct identity by leveraging fine-grained attributes such as his pose and clothing characteristics. In (d), our method retrieves three correct matches, whereas the baseline fails to retrieve any. Despite the presence of two pedestrians in the gallery image, our method is still able to accurately extract robust features. These results highlight the advantage of incorporating richer semantic alignment between vision and language features in complex real-world settings.

\subsection{Computational and commercial costs.}

Computational cost includes text generation and feature compensation. Using Qwen2.5-VL-32B, generating test set texts for CUHK-PEDES, ICFG-PEDES, and RSTPReid took 38, 117, and 12 minutes, respectively, on an I9-13900K CPU with 8 threads. Feature compensation on an RTX 4090 GPU is negligible, much lower than training and inference time. Commercial costs refer to API usage fees for the Qwen2.5-VL-32B, totaling \$22, \$61, and \$7 for the CUHK-PEDES, ICFG-PEDES, and RSTPReid datasets, respectively. In summary, the overall computational and commercial costs are minimal.

\section{Conclusion}

In this paper, we propose a training-free LLM-Collaborative Multi-View Reformulation (MVR) framework to enhance the robustness of text-to-image person retrieval under expression drift and visual ambiguity. By leveraging the generative and reasoning capabilities of large language models, we construct multi-view semantic reformulations that are aggregated in latent space to enrich both textual and visual representations. Extensive experiments demonstrate that our method significantly improves retrieval performance. In future work, we plan to extend our framework to multi-modal reasoning tasks beyond retrieval, further leveraging the complementary strengths of language and vision.

{
    \small
    \bibliographystyle{ieeenat_fullname}
    \bibliography{main}
}

% WARNING: do not forget to delete the supplementary pages from your submission 
\clearpage
\setcounter{page}{1}
\maketitlesupplementary

\section{Computation cost}
For each generation, for DeepSeek-V3 as example, it requires approximately 0.246 TFLOPs per token, making it highly efficient for multi-round generation. In our setting, producing 15 reformulations incurs a total cost equivalent to only a few billion floating-point operations, which can be completed within ~0.5 seconds either on a local GPU or through official APIs. This efficiency ensures that the proposed approach remains real-time applicable in practical systems.

Additionally, integrating IRRA for retrieval introduces merely \textbf{~0.05 TFLOPs} of extra computation, which is negligible compared to the generation cost. Consequently, in a real end-to-end retrieval pipeline, the overall additional latency of our method is around 0.5 seconds, which is comparable to the inference time of the original backbone model and thus has minimal impact on system responsiveness.

\section{Prompts}

\subsection{Our prompt with $P_{key}$}

\begin{minipage}[t]{0.48\textwidth}
\begin{tcolorbox}[
  title=Prompt with $P_{key}$,
  colback=lightgray,
  colframe=gray!70,
  fonttitle=\bfseries,
  boxrule=0.5pt,
  arc=3pt,
  top=1mm, bottom=1mm, left=1mm, right=1mm,
  sharp corners,
]

\lstset{
  basicstyle=\ttfamily,  
  breaklines=true,            
  breakatwhitespace=true,       
  columns=fullflexible,
  keepspaces=true
}

\lstdefinelanguage{Prompt}{
  morekeywords={User, Assistant, System},
  sensitive=true,
  morecomment=[l]{//},
  morestring=[b]",
}

\footnotesize
\begin{lstlisting}[language=Prompt]
System:
  Instructions:
    Suppose you now have a picture of a pedestrian, I will give you a caption and its key words list, your task is to rewrite the caption.
    - Contains every key word must be used, but can change order and replace other words in a similar meaning.
    - Give me 15 different captions and return them in a list. Give me the list without any statement else.
User:
  Caption: {caption}
  Keywords: {keywords}
\end{lstlisting}

\end{tcolorbox}
\end{minipage}

\subsection{Our prompt with $P_{diverse}$}

\begin{minipage}[t]{0.48\textwidth}
\begin{tcolorbox}[
  title=Prompt with $P_{diverse}$,
  colback=lightgray,
  colframe=gray!70,
  fonttitle=\bfseries,
  boxrule=0.5pt,
  arc=3pt,
  top=1mm, bottom=1mm, left=1mm, right=1mm,
  sharp corners,
]

\lstset{
  basicstyle=\ttfamily,    
  breaklines=true,              
  breakatwhitespace=true,      
  columns=fullflexible,
  keepspaces=true
}

\lstdefinelanguage{Prompt}{
  morekeywords={User, Assistant, System},
  sensitive=true,
  morecomment=[l]{//},
  morestring=[b]",
}
\footnotesize

\begin{lstlisting}[language=Prompt]
System:
  Instructions:
    Suppose you now have a picture of a pedestrian, I will give you a caption, your task is to rewrite the caption.
    - The rewrite caption must contain the content mentioned in the original caption.
    - You may use your imagination, but make sure that what you rewrite is not beyound real-world logic.
    - Give me 15 different captions and return them in a list. Give me the list without any statement else.
User:
  Caption: {caption}
\end{lstlisting}

\end{tcolorbox}
\end{minipage}

\section{Examples}
\subsection{Query compensation with $P_{key}$}

\begin{minipage}[t]{0.48\textwidth}
\begin{tcolorbox}[
  title=Example,
  colback=lightgray,
  colframe=gray!70,
  fonttitle=\bfseries,
  boxrule=0.5pt,
  arc=3pt,
  top=1mm, bottom=1mm, left=1mm, right=1mm,
  sharp corners,
]

\lstset{
  basicstyle=\ttfamily,    
  breaklines=true,              
  breakatwhitespace=true,      
  columns=fullflexible,
  keepspaces=true
}

\lstdefinelanguage{Prompt}{
  morekeywords={User, Assistant, System},
  sensitive=true,
  morecomment=[l]{//},
  morestring=[b]",
}
\footnotesize

\begin{lstlisting}[language=Prompt]
Input:
    Caption: This man has short black hair and wears a suit jacket,black trousers with a pair of sneakers.He is looking at left side.
    Keywords: 'man', 'short', 'black', 'hair', 'suit', 'jacket', 'trousers', 'sneakers', 'looking', 'left'.
Output:
[
- A man with short black hair is dressed in a suit jacket and black trousers, paired with sneakers, as he gazes to the left.
- The individual, sporting short black hair, wears a stylish suit jacket, black trousers, and sneakers while glancing toward the left.
- A person with short black hair dons a formal suit jacket, black trousers, and casual sneakers, looking off to the left.
- This gentleman, characterized by his short black hair, is clad in a suit jacket, black trousers, and sneakers, with his gaze directed to the left.
- The man, featuring short black hair, is attired in a suit jacket, black trousers, and sneakers, and appears to be looking toward the left.
...
]
    

\end{lstlisting}

\end{tcolorbox}
\end{minipage}

\subsection{Query compensation with $P_{kdiversity}$}

\begin{minipage}[t]{0.48\textwidth}
\begin{tcolorbox}[
  title=Example,
  colback=lightgray,
  colframe=gray!70,
  fonttitle=\bfseries,
  boxrule=0.5pt,
  arc=3pt,
  top=1mm, bottom=1mm, left=1mm, right=1mm,
  sharp corners,
]

\lstset{
  basicstyle=\ttfamily,    
  breaklines=true,              
  breakatwhitespace=true,      
  columns=fullflexible,
  keepspaces=true
}

\lstdefinelanguage{Prompt}{
  morekeywords={User, Assistant, System},
  sensitive=true,
  morecomment=[l]{//},
  morestring=[b]",
}
\footnotesize

\begin{lstlisting}[language=Prompt]
Input:
    Caption: This man has short black hair and wears a suit jacket,black trousers with a pair of sneakers.He is looking at left side.
Output:
[
- A man with short black hair is dressed in a suit jacket and black trousers, paired with sneakers. He is glancing toward his left side.
- This person has short black hair and is attired in a suit jacket, black trousers, and sneakers. He appears to be looking toward the left side.
- A man sporting short black hair wears a suit jacket, black trousers, and sneakers. He is looking toward the left.
- This person has short black hair and is wearing a suit jacket, black trousers, and sneakers. He is gazing toward the left.
- A man with short black hair is wearing a suit jacket, black trousers, and sneakers. His gaze is directed to the left.
...
]
    

\end{lstlisting}

\end{tcolorbox}
\end{minipage}

\end{document}